\documentclass[letterpaper, 10 pt, conference]{ieeeconf}  

\IEEEoverridecommandlockouts                              

\usepackage{amsmath} 
\usepackage{amssymb}  

\usepackage[pdftex]{graphicx}       
\usepackage[dvipsnames]{xcolor}
\usepackage{algorithm}
\usepackage[noend]{algpseudocode}
\usepackage{float}
\usepackage{siunitx}
\usepackage{hyperref}
\makeatletter
\let\NAT@parse\undefined
\makeatother
\usepackage[numbers]{natbib}

\newcommand{\eref}[1]{(\ref{#1})}
\newcommand{\secref}[1]{Section~\ref{#1}}
\newcommand{\tabref}[1]{Table~\ref{#1}}

\newcommand{\figref}[1]{Fig.~\ref{#1}}

\newcommand{\myparagraph}[1]{\vspace{0.03in}\noindent\textbf{#1}}
\newcommand*{\Cdot}{\raisebox{-0.25ex}{\scalebox{1.75}{$\cdot$}}}

\newboolean{draft-mode}
\setboolean{draft-mode}{true}
\newcommand{\sidenote}[1]{\ifthenelse{\boolean{draft-mode}}{\marginpar{\tiny\raggedright\textsf{\hspace{0pt}#1}}}{}}
\DeclareRobustCommand{\arnote}[1]{\ifthenelse{\boolean{draft-mode}}{\textcolor{blue}{\textbf{AR: #1}}}{}}
\DeclareRobustCommand{\ncdnote}[1]{\ifthenelse{\boolean{draft-mode}}{\textcolor{green}{\textbf{NCD: #1}}}{}}

\title{\LARGE \bf Stable Prehensile Pushing:\\In-Hand Manipulation with Alternating Sticking Contacts}

\author{\authorblockN{Nikhil Chavan-Dafle and Alberto Rodriguez}
  \authorblockA{Department of Mechanical Engineering ---
    Massachusetts Institute of Technology\\
    {\tt\small nikhilcd@mit.edu}, {\tt\small albertor@mit.edu}}
    \thanks{This work was supported by
    NSF award [IIS-1427050] through the
    National Robotics Initiative.}
\includegraphics{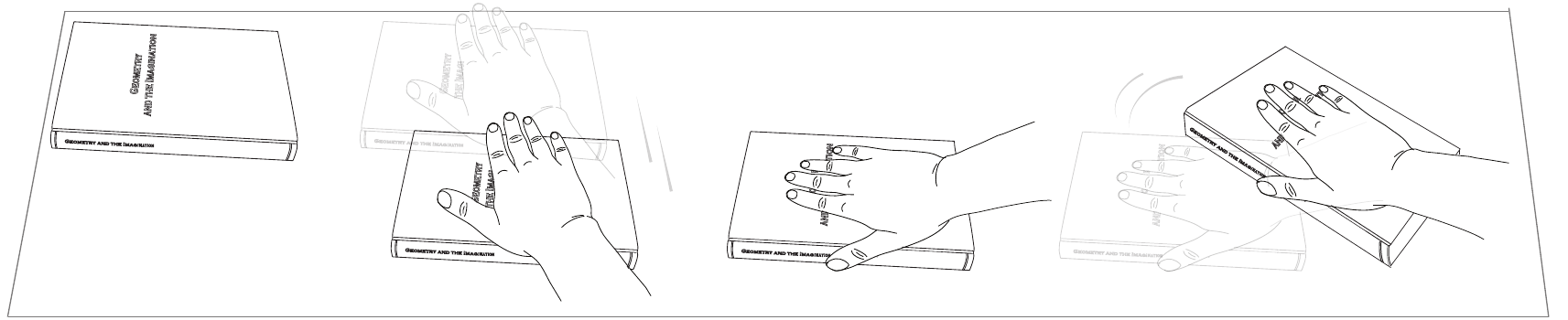}}

\begin{document}

\maketitle
\thispagestyle{empty}
\pagestyle{empty}

\begin{abstract}

This paper presents an approach to in-hand manipulation planning that exploits the mechanics of alternating sticking contact.
Particularly, we consider the problem of manipulating a grasped object using external pushes for which the pusher sticks to the object. %
%
Given the physical properties of the object, frictional coefficients at contacts and a desired regrasp on the object, we propose a sampling-based planning framework that builds a pushing strategy concatenating different feasible stable pushes to achieve the desired regrasp.
An efficient dynamics formulation allows us to plan in-hand manipulations 100-1000 times faster than our previous work which builds upon a complementarity formulation.
Experimental observations for the generated plans show that the object precisely moves in the grasp as expected by the planner.
\\
\href{https://youtu.be/qOTKRJMx6Ho}{\textcolor{RoyalBlue}{Video Summary -- youtu.be/qOTKRJMx6Ho}}
\end{abstract}


\section{Introduction}
\label{sec:intro}
%

\begin{figure*}
\centering
 \includegraphics[width=0.8\textwidth]{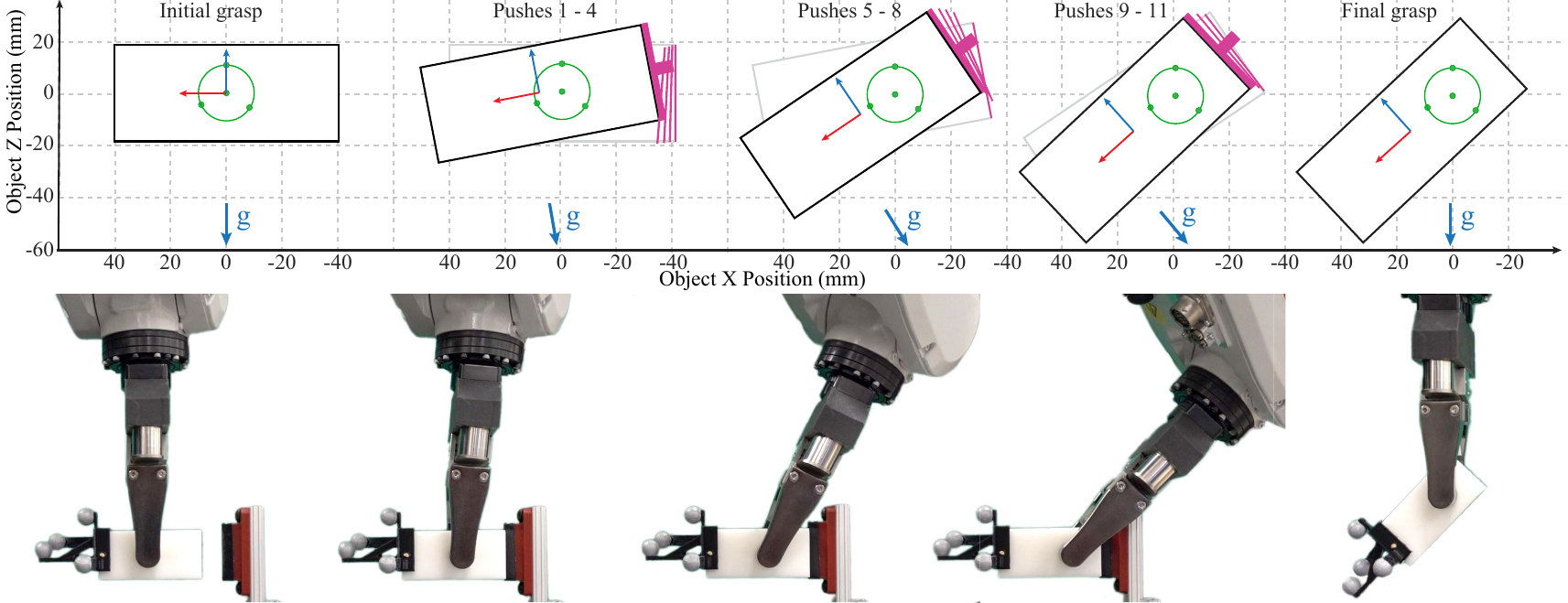}
\caption{An example of stable prehensile pushing. A plastic object is regrasped in a gripper by pushing it against an edge contact in the environment. The upper figure shows a pushing strategy from the planner proposed in this paper. The object motion is shown from a side view; finger contact is a circular patch (in green) and pusher contact is a line contact (in magenta). The lower figure shows the corresponding snapshots from an experimental run.}
    \label{fig:prepush}
\end{figure*} 

This work is inspired from one observation of human manipulation: controlled slip is essential to our dexterous skill, but it is difficult to control the motion of an object in our hands if all contacts slide at the same time.
It is significantly easier when at least one of the contacts involved is securely sticking to the object, albeit that contact might change frequently.
Consider a simpler scenario of a book on top of a table. If both the table and the book were made out of ice, it would be very difficult to control whether the hand slides on the book or the book on the table. With regular coefficients of friction instead, it is simple to manipulate the book with alternating series of motions where either the book sticks to the hand, or the book to the table. It is this idea---forcing and alternating sticking contacts---that we explore in this paper.

This work is also inspired from one technical limitation of robotic in-hand manipulation planning: it is computationally very expensive to find accurate in-hand manipulation plans when all contacts can stick/slide at the same time.
In our previous work, we describe prehensile pushing~\citep{ChavanDafle2015a,ChavanDafle2017} as a particular type of in-hand manipulation that uses pushes against external fixtures to manipulate an object.
In this work, we explore the idea of forcing the external pusher to stick to the object during the pushes, but with a freedom to change its location between the pushes. We will refer to it as \emph{stable prehensile pushing}.
In experimental results, we show that this yields a 100-1000 fold improvement in planning speed, as well as higher accuracy in planned motions.

%

Lynch and Mason~\cite{lynch96} studied the equivalent planar non-prehensile pushing problem where a pusher sticks to an object to control its motion in the plane; they referred to this approach as \emph{stable pushing}. This work follows Lynch and Mason's spirit: ``A model of the mechanics of a task is a resource for the robot, just as actuators and sensors are resources'',
by exploring the role of sticking contact in facilitating fast and robust in-hand manipulation planning. 



The main contribution of this paper is the formulation of the mechanics of stable prehensile pushing, and its application to in-hand manipulation with a parallel-jaw gripper.
We present a planner that works at two levels.
At high-level, an RRT$^*$-based sampling planner efficiently explores the configuration space of grasp poses and builds a tree of grasp poses which can be reached using stable prehensile pushes. 
At low-level a fast dynamics solver for stable prehensile pushing checks if the desired pushes sampled by the high-level planner are feasible given the mechanics of stable prehensile pushing. 
By defining the node cost as a function of the number of pusher switch-overs and exploiting the underlying RRT$^*$ structure, the planner converges to a pushing strategy that forces the object from initial to goal pose while reducing the number of pusher switch-overs. \figref{fig:prepush} shows one of such pushing strategies.

In \secref{sec:examples}, we show multiple examples demonstrating the capability of the proposed planner for manipulating prismatic objects in a parallel-jaw gripper, both in simulation as well as on a robotic system.

In summary, the specific contributions of this paper are:
\begin{itemize}
    \item A model of the mechanics of stable prehensile pushing.
    \item Efficient formulation for testing the feasibility of stable prehensile pushing in a parallel-jaw grasp.
    \item A fast in-hand manipulation planning framework which accounts for the detailed dynamics of frictional contact.
\end{itemize}

\section{Related Work}
\label{sec:related}

In-hand manipulation has motivated the robotics community as early as 1980's. 
%
%
Recent work in robotic manipulation shows rejuvenated interest in in-hand manipulation problem, exploring ideas that diverge from earlier assumptions such as the need for full controllability over the pose of the object using multiple fingers and actuators~\cite{salisbury1982ahf}.

Odhner and Dollar explore the scope of underactuated hands for in-hand manipulation~\cite{Odhner11}. 
Borr\'as and Dollar model a grasp as a parallel mechanism and propose to plan in-hand manipulation as a kinematic problem with compliant joints~\cite{Borras12}.
Sundaralingam and Hermans adopted a similar approach of seeing the in-hand manipulation as a kinematic planning problem but do not constrain the fingers to stick to the object~\cite{Sundaralingam17}. The pure kinematic approach allows them to plan the manipulations in short time; however, experimental results show significant deviations from the final grasp expected by the planner.

Contrary to this approach of changing the object pose in the gripper frame without deliberately changing the locations of finger contacts, some explicitly plan for controlled sliding at fingers.
In our previous work, we presented extrinsic dexterity -- an approach to in-hand manipulation using gravity, dynamic motions of the arm and contacts with the environment~\cite{ChavanDafle2014}. 
%
Holladay et al.~\cite{holladay15} and Vi\~na B. et al.~\cite{kragic_pivoting} demonstrated effective use of dynamic motions and gravity to pivot an object in a parallel-jaw grasp.
Shi et al. show trajectory optimization framework for dynamic in-hand planar manipulations in a pinch grasp~\cite{lynch15}. They explicitly model frictional behaviour at finger contacts with limit surfaces under uniform pressure distribution assumption.

%

Our previous work presented detailed forward as well as inverse dynamics formulations for prehensile pushing \cite{ChavanDafle2015a,ChavanDafle2017}. 
%
We demonstrated that detailed dynamics models with Coulomb friction and maximum energy dissipation principle allow us to predict the motion of the object and forces at contacts realistically~\cite{Kolbert16}. However, solving the resulting mixed nonlinear complementary problem is computationally expensive and limits its use for planning in-hand manipulations~\cite{ChavanDafle2017}.
This compels us to look for a subset of prehensile pushes that could be particularly suited for faster modeling, planning and control.

Interestingly, similar idea is well explored for non-prehensile pushing. Lynch and Mason studied the mechanics of stable pushing and showed its application~\cite{lynch96}.
Zhou and Mason show that planning for stable non-prehensile pushing can be solved as a Dubins car problem \cite{Jiaji17a}.
An assumption of sicking contact allows Huang et al.~\cite{Huang17} to estimate bounds on the object motion even with an unknown support pressure distribution.
%



Prehensile pushing shares conceptual similarities with its non-prehensile equivalent, but is a very different problem from mechanics perspective. 

\myparagraph{Quasi-static vs quasi-dynamic}: In non-prehensile pushing, the quasi-static assumption, with the ellipsoidal limit surface approximation as in \cite{lynch96,Jiaji17b,Hogan16} or with the convex even-degree limit surface model as in \cite{Jiaji17a}, allows a direct mapping of pusher motion to object motion and its consequent benefits for planning and control.

In prehensile pushing, the gravitational force on the object in a grasp is often not perpendicular to the plane of motion. The inclusion of gravity in the dynamics of prehensile pushing makes quasi-static analysis insufficient and calls for a quasi-dynamic or dynamic formulation to account for the effect of gravity on the object motion.

\myparagraph{Changing support locations}: We can view the finger contacts in prehensile pushing as the equivalent to the support surface in non-prehensile pushing. As the fingers move along the object, the effective wrench they produce on the object.
If we were to compute the limit surface from finger contacts or mapping between pusher motion and object motion, we would need to do it at every instance.

On the other hand, for non-prehensile planar pushing, such a wrench set of the support surface remains the same throughout the manipulation and so does the mapping between the pusher motion and the slider motion.

So, the work on non-prehensile pushing which relies on the quasi-static assumption and consequent benefits for planning and control does not extend directly to prehensile pushing, not even to planar reconfigurations in a pinch grasp.

\section{Framework for In-hand Manipulation Planning}
\label{sec:formulation}

This paper presents a planner for in-hand manipulations. We demonstrate that by limiting to the set of object motions for which at least one of the contacts sticks to the object, we can speed up the planning for in-hand manipulations significantly.

In our implementation of stable prehensile pushing, we execute the pushes using the environment. However, one could implement similar pushes using a second robot arm or different fingers of a multi-finger gripper.
As a general approach, we assume that an object is grasped in a gripper which is fixed in the world. A virtual pusher with full control and freedom to change it's location on the object executes the external pushes. 

Our algorithm starts with the following information about the physical properties of the manipulation system:
\begin{itemize}
    \item[$\Cdot$] Object geometry and mass.
    \item[$\Cdot$] Initial and goal pose of an object in a grasp, specified by the locations and geometries of each fingers contacts.
    \item[$\Cdot$] Gripping force.
    \item[$\Cdot$] Discrete set of pusher contacts, specified by their locations and geometries.
    \item[$\Cdot$] Coefficient of friction at all contacts.
\end{itemize} 

An efficient dynamics formulation of stable prehensile pushing (\secref{sec:stable_prepush_mechanics}) allows us to query if certain prehensile push is a stable push or not. This fuels a sampling-based planner which then rapidly explores the configuration space of possible grasps and builds a tree of grasp poses reachable with stable prehensile pushes.

The planner iterates through the following steps:
\begin{itemize}
    \item[i.] Sample a random object pose in a grasp.
    \item[ii.] Check if moving toward the sampled pose satisfies a ``benefit" criteria. If not, return to step i.
    \item[iii.] Check if it's possible for any of the pushers to force the object towards the sampled pose with a \textit{stable push}. If not possible, return to step i.
    \item[iv.] Check for other ways to reach the new pose from the surrounding nodes in the tree with a lower cost.
    \item[v.] Iterate until the goal pose in the grasp is reached within a given resolution and cost threshold.
\end{itemize}

\section{Mechanics of Stable Prehensile Pushing}
\label{sec:dynamics}

Prehensile pushing is an intricate manipulation problem where geometries, physical properties, frictional parameters of the gripper, the grasped object, and the environment all play an important role in determining the resultant motion of the object in the grasp when pushed. 



\subsection{Contact Modeling}
\label{sec:contact}

\begin{figure}
\centering
 \includegraphics[scale=1.35]{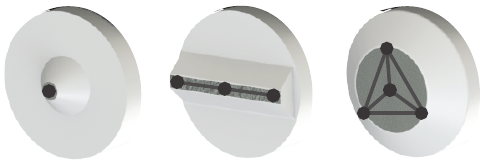}
\caption{Different contact geometries: point, line and circular patch, modeled as sets of rigidly connected point contacts. Figure adapted from \cite{ChavanDafle2017}.}
    \label{fig:contact_shape}
    
\vspace{-3mm}
\end{figure} 
We model interaction between gripper, grasped object, and the pusher, by discretizing complex surface/line contacts into arrays of hard frictional point contacts as shown in \figref{fig:contact_shape}.
We define a point contact between two bodies by a local coordinate frame with, $\hat{\boldsymbol{n}}$ normal to the contact plane and and $\hat{\boldsymbol{t}}$ and $\hat{\boldsymbol{o}}$, two orthonormal direction vectors, in the contact plane.
Let $\boldsymbol{f}=[f_{n}, f_{t},f_{o}]^\top$ and $\boldsymbol{v}=[v_n,v_t,v_o]^\top$ be a net force and a relative velocity at a contact in the local contact frame.
For a given coefficient of friction ($\mu$), Coulomb's friction cone is defined as:
\begin{equation}
\label{eq:fricCone}
FC=\{f_{n}\hat{\boldsymbol{n}} +
f_{t}\hat{\boldsymbol{t}} + f_{o}\hat{\boldsymbol{o}}\
| \ f_{n} \geq 0,\ f_{t}^2+f_{o}^2 \leq \mu f_{n}^2\}
\end{equation}
By Coulomb's law, when a contact slides, the contact force is on the boundary of the friction cone and the direction of the friction force is opposite to that of the sliding velocity at the contact.
We can write these constraints using complementarity and nonlinear equations:
\begin{equation}
\label{eq:compfricbound}
[(\mu f_n)^2 - f_t^2 -f_o^2]\sqrt{v_t^2+v_o^2}=0
\end{equation}
\vspace{-3mm}
\begin{equation}
\label{eq:quad_dissipation}
  \mu f_{n}v_{i}+f_{i}\sqrt{v_t^2+v_o^2}=0 \hspace{1cm} i=t, o
\end{equation}

We need another complementarity constraint to make sure that a two bodies exert forces only when they are in contact.
\begin{equation}
\label{eq:unilateral}
  v_{n}\cdot f_{n}=0,\  v_{n}\geq 0,\
  f_{n}\geq 0
\end{equation}

Note that both the complementarity constraints \eref{eq:compfricbound} and \eref{eq:unilateral} are responsible to capture the appropriate force-motion interrelationship based on different contact modes. If contact modes are known a priori, we can constrain forces only with nonlinear constraints \eref{eq:fricCone} and \eref{eq:quad_dissipation}.

\subsection{Dynamics of Prehensile Pushing}
\label{sec:prepush_dynamics}

We model prehensile pushing as a general rigid body dynamics problem where object motions are governed by kinetic and kinematic constraints.

\myparagraph{Newton Euler Equation}:
\label{sec:newton}
Let $\mathbf{G_i}$ maps the local contact forces at a contact $i$ to a wrench in the object frame. $\mathbf{G}_\textnormal{finger}$ is diagonal concatenation of $\mathbf{G_i}$'s for all the finger contacts on the object and similarly $\mathbf{G}_\textnormal{pusher}$. 
In the quasi-dynamic framework, for a single time step with zero initial velocity of the object, we can write time-integrated Newton's law for an object with mass $m$ and generalized inertia matrix $\mathbf{M}$ as:
\begin{equation}
\label{eq:newton_vel}
\mathbf{G_\textnormal{finger}}\cdot \boldsymbol{P_\textnormal{finger}} + \mathbf{G_\textnormal{pusher}}\cdot \boldsymbol{P_\textnormal{pusher}}+ \vec{P}_{mg} = \mathbf{M}\cdot {\vec{v}}_{\textnormal{obj}}
\end{equation}

\noindent where $\boldsymbol{P_\textnormal{finger}}$ and $\boldsymbol{P_\textnormal{pusher}}$ are arrays of impulses equivalent to all the finger and pusher contact forces resp. $\vec{P}_{mg}$ is the gravitational impulse and ${\vec{v}}_{\textnormal{obj}}$ is the resultant object velocity.

\myparagraph{Rigid Body Motion Constraints}:
\label{sec:rigid_body}
Let $\mathbf{J}_\textnormal{pusher}$ be a jacobian matrix that maps the velocities of pusher actuators ($\boldsymbol{\dot{\theta}}_\textnormal{pusher}$) to the input velocities at all the pusher contacts in the local contact frames. Let $\boldsymbol{V}_\textnormal{pusher}$ be an array collecting relative velocities at all the pusher contacts.
%
\begin{equation}
\label{eq:pusher_vel}
\boldsymbol{V_\textnormal{pusher}} = \mathbf{G}^\top_\textnormal{pusher}\cdot\vec{v}_{\textnormal{obj}} - \mathbf{J_\textnormal{pusher}}\cdot \boldsymbol{\dot{\theta}}_\textnormal{pusher}
\end{equation}
Equivalently, for finger contacts,
\begin{equation}
\label{eq:finger_vel}
\boldsymbol{V_\textnormal{finger}} = \mathbf{G}^\top_\textnormal{finger}\cdot\vec{v}_{\textnormal{obj}} - \mathbf{J_\textnormal{finger}}\cdot \boldsymbol{\dot{\theta}}_\textnormal{finger}
\end{equation}

\noindent In our prehensile pushing framework, finger contacts respond passively, while the object is being actively pushed by the pusher, i.e., $\boldsymbol{\dot{\theta}}_\textnormal{finger}=\boldsymbol{0}$.

Forward dynamics for prehensile pushing as presented in \cite{ChavanDafle2015a} refers to finding the object motion, given the finger and pusher motions.
The inverse dynamics problem similarly refers to finding the pusher motion, given a desired object motion in the grasp \cite{ChavanDafle2017}.
Note that for solving forward and inverse dynamics problems, the relative motions and forces at all the contacts need to be solved simultaneously.

\subsection{Dynamics of Stable Prehensile Pushing}
\label{sec:stable_prepush_mechanics}

Stable prehensile pushes are those pushes for which the pusher sticks to the object.
If the pusher sticks to the object, contact modes at all the contacts are fully defined during the push. The knowledge of object motion or pusher actuator motion is sufficient to find the relative velocities at all the contacts from rigid body motion constraints \eref{eq:pusher_vel} and \eref{eq:finger_vel}.
%
Consequently, for stable prehensile pushing, solving the forward or inverse dynamics boils down to a feasibility check.
For a sampled object motion or pusher motion, we need to check if there are forces at the contacts that satisfy the Newton Euler equation \eref{eq:newton_vel} while satisfying the constraint \eref{eq:fricCone} or \eref{eq:quad_dissipation} depending on the relative motions at the contacts.
This general problem can be solved as a nonlinear constraint satisfaction problem. In this paper, motivated by fast in-hand manipulation planning, we focus on an application of this framework for pinch grasp manipulations.


For pinch grasps or parallel jaw grasp, manipulations are constrained in a 2D plane. The sum of the normal forces at each finger contact must satisfy the force balance along the perpendicular to the manipulation plane. It is also common to assume that pressure distribution at fingers will be uniform for these manipulations~\cite{lynch15}. With these constraints, normal force at all the constituent finger contacts can be computed given the gripping force.
For a given object motion or pusher motion, depending on the contact mode at every finger, we can estimate either a unique frictional wrench or a possible set of frictional wrenches that finger contacts will resist the desired object/pusher motion with. 

\subsection{Dynamics of Stable Prehensile Pushing - when all finger contacts slide}
\label{sec:all_slide}
When a desired object/pusher motion is such that all the finger constituent contacts slide on the object, frictional force at each of them is uniquely determined by the maximum energy dissipation principle \eref{eq:quad_dissipation}.
We can rearrange Newton Euler equation \eref{eq:newton_vel} as:
\begin{equation}
\label{eq:newton_allslide}
\mathbf{G_\textnormal{finger}}\cdot \boldsymbol{P_\textnormal{finger}} + \vec{P}_{mg} - \mathbf{M}\cdot {\vec{v}}_{\textnormal{obj}}=-\mathbf{G_\textnormal{pusher}}\cdot \boldsymbol{P_\textnormal{pusher}}
\end{equation}

Note that all the terms on the left side of the equation \eref{eq:newton_allslide} are known and constitute a net motion wrench -- a force in the grasp plane and a torque perpendicular to the grasp plane, i.e.,  about the finger axis. For a push to be a stable push, the pusher has to provide this wrench while sticking to the object.

\begin{figure}
\centering
 \includegraphics[scale=0.9]{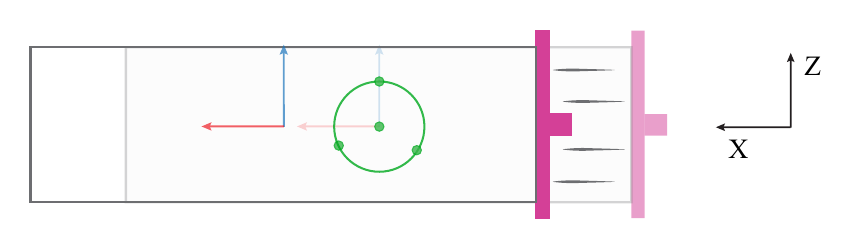}
\caption{A schematic of a simple planar prehensile pushing instance where all finger contacts slide.}
    \label{fig:all_slide}
    \vspace{-5mm}
\end{figure} 
\figref{fig:all_slide} shows a schematic of a simple planar prehensile pushing instance where all the constituent finger contacts slide. All the finger constituent contacts slide linearly along $X$; they generate frictional impulse along $-X$. Object momentum, gravitational impulse and contact impulses from fingers add up to a net motion wrench -- a force in the grasp plane ($XZ$ plane).
%
This is a feasible stable prehensile push only if there exists an impulse $\boldsymbol{P_\textnormal{pusher}}$ that satisfies constraint \eref{eq:newton_allslide} and $p_{t}^2+p_{o}^2 < \mu p_{n}^2$ for all the pusher contacts. In other words, if the motion wrench required is inside the generalized friction cone of the pusher contact, stable prehensile pushing is possible.

\myparagraph{Generalized Friction Cone for a Pusher}: generalized friction cone for a pusher is a wrench cone that the pusher can generate in the object frame as a consequence of forces at the pusher contacts~\cite{Erdmann94}. For multi-point pusher, the generalized friction cone can be computed as a vector sum of generalized friction cones for constituent point contacts~\cite{Erdmann93, lynch96}.

\begin{figure}
\centering
 \includegraphics[scale=0.9]{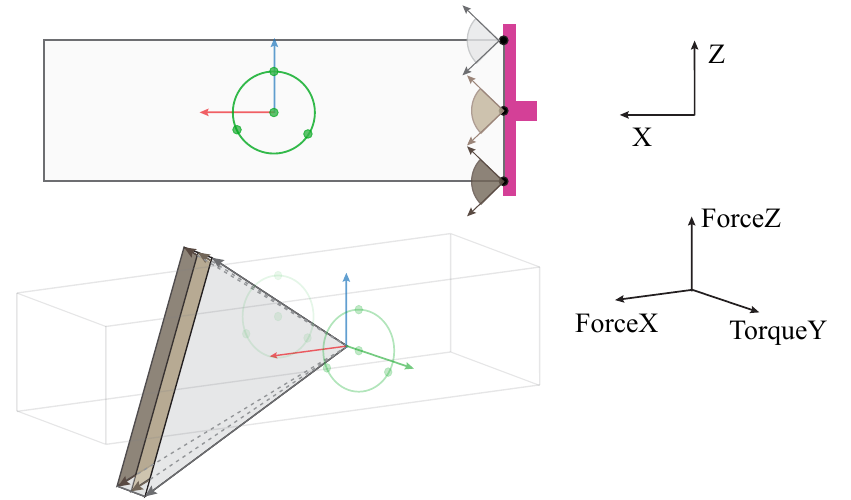}
\caption{Generalized friction cone for a line pusher in example in \figref{fig:all_slide}. Note the contribution of each constituent point contact.}
    \label{fig:gen_cone}
    \vspace{-2mm}
\end{figure}

For the example in \figref{fig:all_slide}, the line pusher is modelled with three points. By mapping forces at each contact through their respective $G_i$, we can find the generalized friction cone for all the pusher constituent contacts. Finally, the vector sum of the wrench cones for all the contacts provide the generalized friction cone for the line pusher as shown in~\figref{fig:gen_cone}. 

\begin{figure}
\centering
 \includegraphics[scale=0.9]{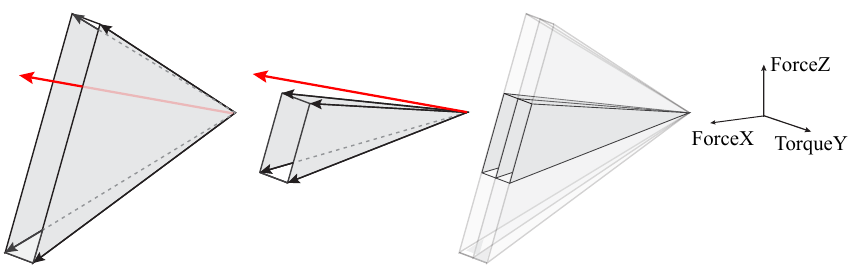}
\caption{Feasibility check for a stable push shown in \figref{fig:all_slide}. Figure on the right shows how the generalized friction cone changes when friction at the pusher changes from 0.2 to 0.6. Note that the motion in \figref{fig:all_slide} is a stable push for high friction pusher, but not for low friction pusher.}
\label{fig:allslide_check}
\vspace{-3mm}
\end{figure}

\myparagraph{Feasibility Check}:
Evaluating if the required motion wrench can be provided by the pusher is now equivalent to checking if the wrench falls inside the generalized friction cone of the pusher.
We do it by projecting the wrench on the face normal representation of the generalized friction cone \cite{Hirai91}. If all the projections are negative, the motion wrench falls inside the wrench cone, not otherwise.
\figref{fig:allslide_check} shows the feasibility of the motion in \figref{fig:all_slide} for different pushers.


\subsection{Dynamics of Stable Prehensile Pushing - when at least one of the finger contacts sticks}
\label{sec:stick_slide}

When one of the finger contacts sticks, frictional forces at that contact are not uniquely defined; the only constraint we have is that they have to be inside the friction cone.


Further rearranging \eref{eq:newton_allslide}, we can see that, a given prehensile push of this type can be stable prehensile push only if any of the wrenches in the wrench set formed by forces at sticking finger contacts and the required motion wrench computed from rest of the terms on the right side of equation \eref{eq:newton_stickslide} falls in the generalized friction cone of the pusher.
\vspace{-1mm}
\begin{equation}
\label{eq:newton_stickslide}
\begin{split}
\mathbf{G_\textnormal{slidingfingers}}\cdot \boldsymbol{P_\textnormal{slidingfinger}} + \vec{P}_{mg} - \mathbf{M}\cdot {\vec{v}}_{\textnormal{obj}} \hspace{1.7cm}\\
+ \  \mathbf{G_\textnormal{stickingfingers}}\cdot \boldsymbol{P_\textnormal{stickingfingers}} =
- \mathbf{G_\textnormal{pusher}}\cdot \boldsymbol{P_\textnormal{pusher}}
\end{split}
\end{equation}

\begin{figure}
\centering
 \includegraphics[scale=0.9]{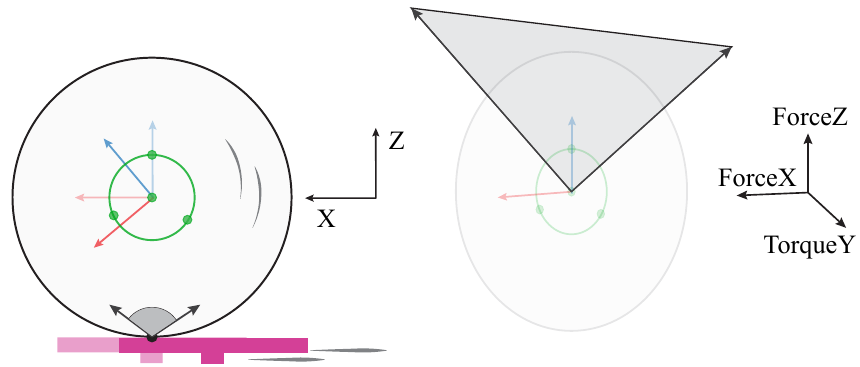}
\caption{Schematic of a prehensile rolling manipulation where center constituent contact on the finger sticks while rest of them slide. Figure on the right shows generalized friction cone for the point pusher.}
    \label{fig:stick_case}
    \vspace{-2mm}
\end{figure}

\begin{figure}
\centering
 \includegraphics[scale=0.9]{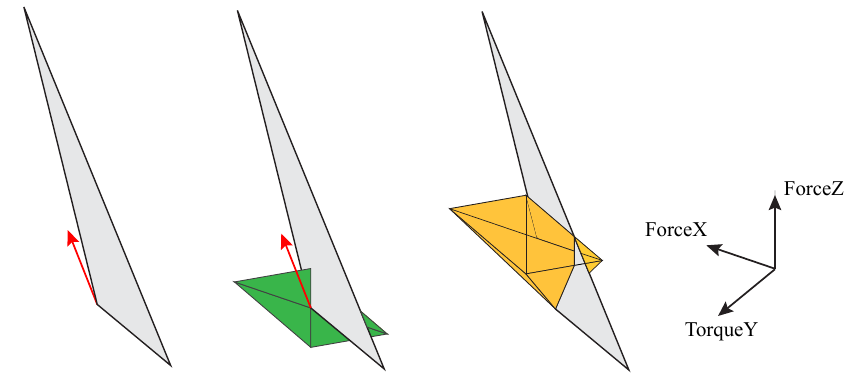}
\caption{Feasibility check for a stable push shown in \figref{fig:stick_case}. From left to right: 1) the given push would be deemed not stable as motion wrench falls outside the generalized friction cone of the point pusher, 2) Green polyhedron shows possible friction wrench that sticking finger contact can provide, 3) The wrench cone (shown in yellow) formed by vector sum of motion wrench (shown in red) and the wrench cone from sicking finger contact (shown in green) intersects with the generalized friction cone of the pusher indicating that the motion in \figref{fig:stick_case}, is a stable push.}
\label{fig:sickslide_check}
\vspace{-2mm}
\end{figure}

\figref{fig:stick_case} shows a schematic of another simple prehensile pushing example - rotating a circular disc using a point contact with the ground. 
Note that for pure rolling motion, all the finger constituent contacts except the one at the center will side. Considering forces only at sliding contacts we get a pure frictional torque about $Y$ axis. Adding it to the contribution from gravity impulse and object momentum gives us a net wrench which has force component along $Z$ and torque about $Y$. 
To bring in the possible contribution from the sticking finger contact, we take a vector sum of the motion wrench computed above and the generalized friction cone from sticking finger contact as shown in \figref{fig:sickslide_check}. Considering a polyhedral approximation to the friction cone at sticking contacts with a sufficiently large number of facets, provides an efficient way to generate the generalized friction cone without compromising on the model accuracy. In \figref{fig:sickslide_check} we show polyhedral friction cone with only 4 facets (shown in green) for visualization simplicity, but for actual computations we use 16 facets.

\myparagraph{Feasibility Check}:
The feasibility check in this case boils down to checking if the polyhedral wrench sets on either sides of the equation \eref{eq:newton_stickslide} intersect or not.
This can be efficiently done using the standard technique of taking union of their face normals and then verifying that the set formed by the union is not empty.


Note that in the above example if we were to neglect the forces at sticking contacts, the given push will not be a feasible prehensile push as seen in \figref{fig:sickslide_check}. The pusher contact can not generate any torque about Y unless it applies a force along $X$. It is only when we consider the frictional force at the sticking finger contact balancing this pusher force along $X$, the prehensile push becomes feasible. This confirms that during a pure rolling motion, the contact on the ground instantaneously sticks, which is consistent with the definition of pure rolling.


\section{Planning for Stable Pushing Strategies}
\label{sec:planning}
Underactuation in the dynamics of pushing and the workspace constraints for in-hand manipulation foils the greedy approach of pushing always toward the desired goal.
For general in-hand manipulations, we expect the discrete pusher contact switch-overs to play a crucial role to force the object to a desired pose. 
These constraints call for a long-horizon planning framework that can reason about discrete pusher changes and also allow the pushing strategy to deviate from the goal momentarily if it's necessary to get the object eventually to a goal grasp.
Our previous work demonstrated effective use of a sampling-based planning architecture for planning in-hand manipulations with prehensile pushes~\cite{ChavanDafle2017}. We follow a very similar framework in this paper. 

Our planner works at two levels. 
A high level planner efficiently explores the configuration space and builds a tree of grasp poses that are reachable using prehensile pushing. The low-level dynamics checks the dynamics feasibility of the pushes sampled by the high-level planer.

\subsection{Low-Level: Dynamics Feasibility Check}
The two cases considered in \secref{sec:stable_prepush_mechanics} cover all the planar manipulations in a pinch grasp. Using, the feasibility checks explained in \secref{sec:stable_prepush_mechanics} we evaluate if the desired object motion in the grasp is possible with stable prehensile push using any of the pushers given. This computation can be done efficiently, in less than \SI{100}{\micro\second} in our MATLAB implementation . In contrast, our optimization based formulation in \cite{ChavanDafle2017} to solve the inverse dynamics of instantaneous prehensile pushing takes 2-3 seconds.

As we limit the possible prehensile pushes to only stable prehensile pushes, the locations of the pushers specified while initiating the planner do not change in the object frame. Generalized friction cones for pushers computed at the beginning stay the same throughout the manipulation, adding to the time efficiency.

\subsection{High-Level: Long Horizon Planner}
\begin{algorithm}
  \caption{: In-Hand Manipulation Planner}\label{alg:full_planner}
  $  \textbf{input}: q_{init}, q_{goal}$ \par
  $  \textbf{output}:$ {tree} $\ \mathcal{T}$
  \begin{algorithmic}
  \State $\mathcal{T}\gets \textrm{initialize tree}(q_{init})$
      \While{$q_{goal} \notin \mathcal{T}$ \textbf{or} cost($q_{goal}) > \textrm{cost threshold} $}
        \State $q_{rand}\gets \textrm{sample random configuration}(\mathcal{C})$
        
        \State $q_{parent}\gets \textrm{find nearest neighbor}(\mathcal{T},q_{rand})$
        \While{$q_{rand} \notin \mathcal{T}$}
        \State $q_{new}\gets \textrm{take unit step}(q_{parent},q_{rand})$
        
        \If{\textrm{transition test}$(q_{parent},q_{new},\mathcal{T})$ \textbf{and} \\ \hspace{2cm}\textrm{grasp maintained}$(q_{new})$} 
            
            \State $\textrm{stable feasible} \gets \textrm{stable check}(q_{parent},q_{new})$
            
            \If{stable feasible}
            \State $q\mbox{*}_{parent}\gets \textrm{optimEdge}(\mathcal{T},q_{new},q_{parent})$
            
            \State $\textrm{add new node}(\mathcal{T},q_{new})$
            
            \State $\textrm{add new edge}(q\mbox{*}_{parent},q_{new})$
            
            \State $\textrm{rewire tree}(\mathcal{T},q_{new},q\mbox{*}_{parent})$
            
            \State $q_{parent}\gets q_{new}$
            \Else            \State \textbf{break}
            \EndIf
            \Else
            \State \textbf{break}
        \EndIf
        \EndWhile
    \EndWhile
  \end{algorithmic}
\end{algorithm}

We use Transition-based RRT$^*$ optimal sampling framework~\cite{trrt_star} for high-level planning .

\myparagraph{For controlled exploration} of the configuration space, we define the configuration cost as a distance from goal. The transition test loosely confines the stochastic exploration towards the goal grasp, while allowing the flexibility to explore in other directions when necessary to get the object finally to the goal pose.

\myparagraph{For optimal connections} in the tree, we define the node cost as a weighted sum of its distance away from the goal and the number of pusher switch-overs. The underlying RRT$^*$ architecture makes and rewires the connections in the tree such that the number of pusher switch-overs needed to push the object to a desired pose is minimized. This avoids the unnecessary noise that may get introduced into the manipulation whenever pusher contact is changed.
More details on the configuration and node cost definitions and on the procedure for transition test can be found in our previous work \cite{ChavanDafle2017}.

Algorithm \ref{alg:full_planner} shows our in-hand manipulation planner.
Let $q$ denote a configuration of an object, i.e., the pose of the object in a gripper frame which is fixed in the world.
For the scope of this paper we are interested in planar manipulations in a pinch grasp, so the configuration space is $[X, Z, \theta_y] \in {\rm I\!R}^3$, i.e., the object can translate in the grasp plane ($XZ$) and rotate around a perpendicular ($Y$) to the grasp plane.

Let $q_{init}$ and $q_{goal}$ be an initial and desired pose of the object respectively. 
The planner initiates a tree $\mathcal{T}$ with $q_{init}$. 
While the desired object pose is not reached within some cost threshold, a random configuration ($q_{rand}$) is sampled and nearest configuration to $q_{rand}$ in the tree $\mathcal{T}$ is found. 
The object is tried to be pushed towards $q_{rand}$ as long as the unit step motion of the object satisfies the transitions test and does not move the object outside the grasp.
Using the low-level dynamics feasibility check, it is evaluated if every unit step motion towards $q_{rand}$ is stable prehensile push or not. Only the stable prehensile pushes are added to the tree.

At every unit step propagation of the tree, \textit{optimEdge} routine makes sure that the connection to $q_{new}$ is made such that it's cost is locally minimized.
Similarly, \textit{rewire tree} routine checks the nodes around $q_{new}$ for restructuring the connections in the tree to reduce their cost. Both these routines are directly adopted from RRT$^*$ architecture originally proposed in~\cite{rrt_star} and also used in our previous work \cite{ChavanDafle2017}. 

The low-level dynamics check and the high-level planner work together to build a tree of grasp poses connected with stable prehensile pushes. Note that the planner continues to make and rebuild new connections in the tree until it finds a stable prehensile pushing strategy to force the object to the goal region with minimal number of pusher switch-overs, less than some predefined threshold.

\section{Examples and Observations}
\label{sec:examples}

\begin{table}[b]
  \caption{Physical properties of the objects used.}
  \label{tab:objects}
	\centering
	\begin{tabular}{|l|l|c|r|}
         \hline
          \textbf{Shape} & \textbf{Material} & \textbf{Dim [L, B, H]} (mm) & \textbf{Mass} (g) \\ \hline
          square prism  & Al 6061 & 100, 25, 25 & 202\\ \hline  
          rectangular prism  & Delrin & 80, 25, 38 & 130\\ \hline
          T-shaped & ABS  & 70, 25, 50 & 62\\ \hline
    \end{tabular}
\end{table}

We evaluate the performance of our planner for manipulations performed with a parallel jaw gripper and different objects listed in \tabref{tab:objects}. We validate the planned pushing strategies on a manipulation platform instrumented with an industrial robot arm, a parallel-jaw gripper with force control, features in the environment that act as pushers, and a Vicon system for object tracking. 

\subsection{Example manipulations and experimental results}
We consider a parallel-jaw gripper with flat circular finger contacts as seen in the simulation figures. 
The planner is initiated with pusher contacts on either sides of the object and under the object. 
All the pusher contacts are line/edge contacts.
For planing and simulation, we used a computer with Intel Core i7 2.8 GHz processor and MATLAB R2017a. 
Initial pose of the object is treated as $[0,0,0]$ and goal poses for different examples are listed in \tabref{tab:timing}. The planning time shown in \tabref{tab:timing} is in seconds and is median time over 10 planning trials.

\subsubsection{Regrasping an object offset to the center}
The goal in this seemingly simple example is to regrasp the square prism horizontally 20 mm offset from the center. The gripper force and frictional parameters at fingers and features in the environment are chosen such that pushing the object horizontally will not be a valid solution and the object will also slide downwards by a few millimeters~\cite{Kolbert16}. 

\begin{figure}
\centering
 \includegraphics[scale=0.69]{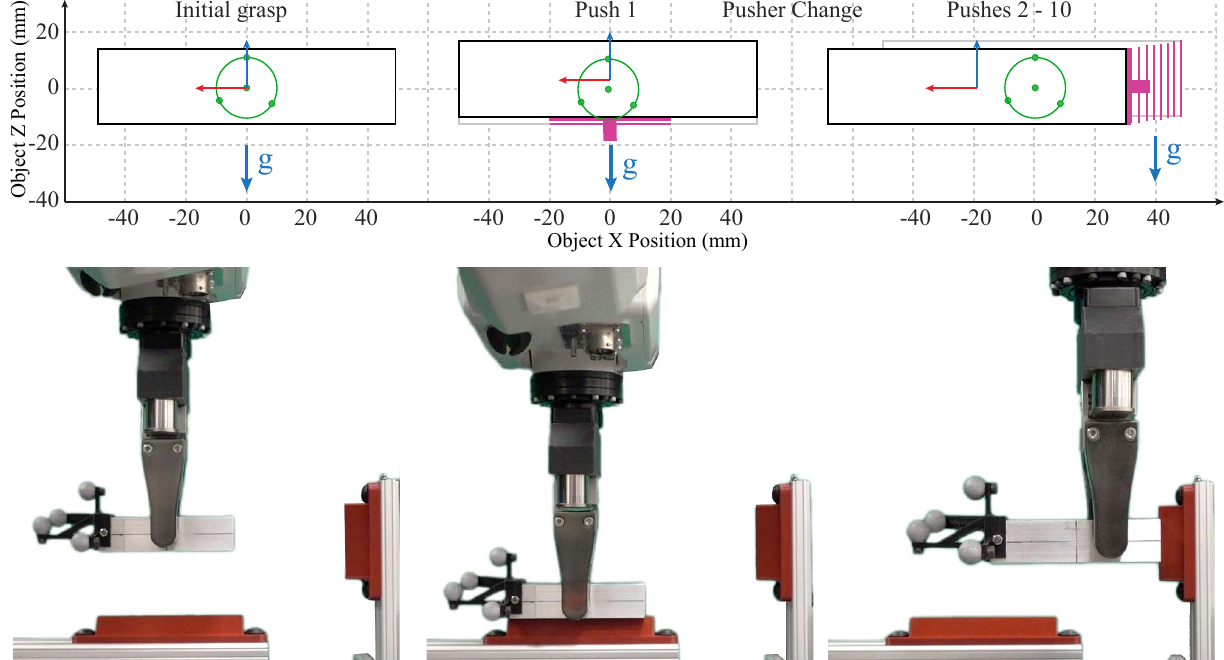}
\caption{Simulated motion of the object and snapshots of the experimental run for a pushing strategy that the planner converged to push the object straight in the grasp using low coefficient pushers.}
    \label{fig:linpush}
    \vspace{-4mm}
\end{figure}

With these constraints, our planner comes up with a plan which first strategically slides the fingers down using the bottom contact and then uses the pusher on the side to drag the fingers up and along the length of the object as seen in~\figref{fig:linpush}. This strategy is similar to the one we observed in our previous work~\cite{ChavanDafle2017} too, however the planning time required by the current planner is more than 200 times smaller as shown in Table \ref{tab:timing}. 

If we increase the coefficient of fiction for the contacts in the environment, we can increase the scope of stable prehensile pushing to wider set of regrasps. For the following manipulations, we use pusher contacts with rubber coating which provides approximately 0.6 coefficient of friction.

\begin{figure}[t]
\centering
 \includegraphics[scale=0.815]{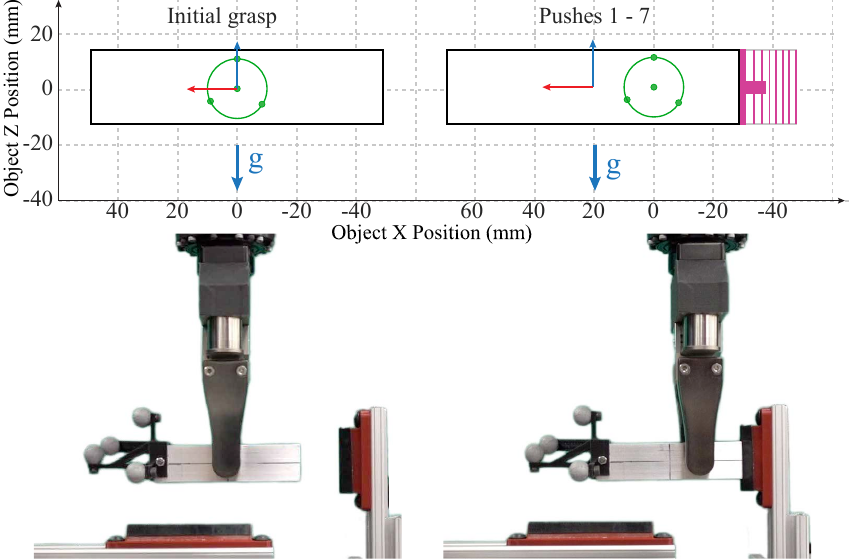}
\caption{Simulated motion of the object and snapshots of the experimental run for a pushing strategy using high coefficient pushers.}
    \label{fig:linpush_highf}
\end{figure}

With high friction at the pushers, for the example considered above, our planner converges to a plan where it uses only the side pusher to push the object straight in the gasp as shown in~\figref{fig:linpush_highf}.
Experimental runs validated the pushing strategies generated by the planner for different pusher contacts. For 10 experimental trials of the pushing strategy shown in \figref{fig:linpush}, we get a deviation of [$-0.08$ to $0.02$ mm, $-0.05$ to $0.24$ mm, $-0.09$ to $-0.03$ degree] in [$X, Z, \theta_y$] in the final expected object pose in the grasp. 

\subsubsection{Pivoting an object in the grasp}
In this example we try to pivot the square prism in the grasp. The planner converges to a strategy exactly same as the one we found in our earlier work~\cite{ChavanDafle2017}, where the side pusher is rotated about the center of the object in the hand.
Note that, for pivoting, center points on the finger contacts do not slide similar to the rolling example in \figref{fig:stick_case}. The contribution of the sticking contact is essential here as discussed in \ref{sec:stick_slide}.

\subsubsection{General manipulation in [$X, Z, \theta_y$]}
This example demonstrates large in-hand manipulation for the rectangular prism. The planner quickly finds a solution which often involves multiple pusher changes, but very soon converges to a strategy which does not require any pusher switch-over, for example the one shown in \figref{fig:prepush}. A few times, the planner directly found the strategy with no pusher switch-over. 


For 10 experimental trails of the pushing strategy shown in \figref{fig:prepush}, we observed a deviation of [$-0.04$ to $0.07$ mm, $-0.37$ to $0.09$ mm, $-0.13$ to $0.48$ degree] in [$X, Z, \theta_y$] in the final pose of the object in the grasp compared to the pose simulated by the planner.

\begin{table}
  \caption{Planning times (in seconds) for different manipulations.}
  \vspace{-2mm}
  \label{tab:timing}
	\centering
	\begin{tabular}{|l|c|r|r|}
         \hline
          \textbf{Manipulation} & \textbf{Goal [$X, Z, \theta_y $]} & \textbf{Planning}  & \textbf{Planning}\\
           & [mm, mm, deg] & \textbf{Time}[stable]  & \textbf{Time}~\cite{ChavanDafle2017} \\
          \hline
          Horz. offset (low $\mu$)  & 20, 0, 0 & 2.83 & 592.8\\ \hline  
          Pivoting  & 0, 0, 90 & 1.6 & 128.4\\ \hline
          Large Manipulation & 15, -13, 45  & 2.54 & 17684\\ \hline
          T-shaped & 25, 17.7, 0  & 0.82 & 32657\\ \hline
    \end{tabular}
    \vspace{-3mm}
\end{table}

\subsubsection{Manipulation with a complex-shaped object}
This example shows manipulation of a T-shaped object. Pushing the object directly towards the goal would lead to loosing the grasp on the object, so the role of the long horizon planning is essential for regrasping such concave objects.
We initiate the planner with four pushers. One on either sides of the object, one on the web, one on the flange, and two under the object, one on the web and one on the flange.

\begin{figure}[t]
\centering
 \includegraphics[scale=0.85]{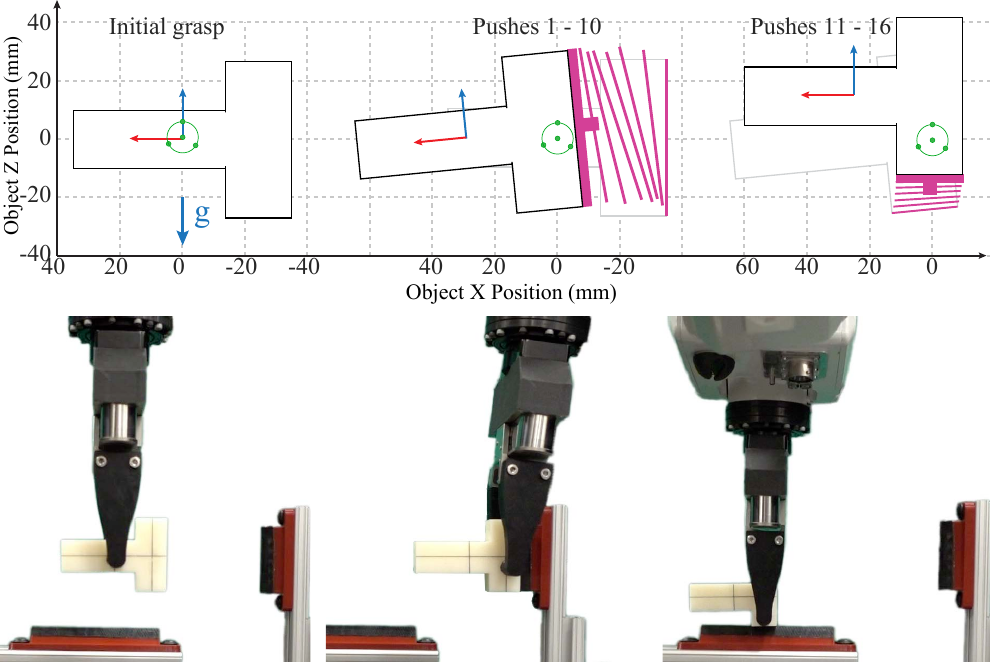}
\caption{A pushing strategy for manipulating the T-shaped object}
    \label{fig:tpush}
    \vspace{-3mm}
\end{figure}

\figref{fig:tpush} shows the strategy our planner came up with and consequent motion of the object in the grasp.
Note that the planner uses the pusher under the flange to correct the small overshoot in the X direction as it pushes the object up. 
%
The planner is efficient for these complex regraps too.

\subsection{Observations}
\myparagraph{Fast planning}:
The computationally efficient formulation of dynamics of prehensile prehensile pushing allows the planner to rapidly explore the configuration space of different grasps and connect them with stable pushes. As shown in \tabref{tab:timing}, we see 100 to 1000 times increase in the planning speed compared to our previous work on planning for prehensile pushing~\cite{ChavanDafle2017}.
Though theoretically the set of object motions that can be generated using stable prehensile pushing is smaller than that using general prehensile pushing~\cite{ChavanDafle2017}, we do not find it limiting, especially with high coefficient of friction at pushers. The large advantage of time efficiency outweighs the reduction in dexterity.

\myparagraph{Observable actions}:
The experimental runs confirmed that the pushing strategies generated by our planner result into the motions where the object sticks to the pusher. This can also be seen in the figures with the experimental runs and also in the attached video~\footnote{\href{https://youtu.be/qOTKRJMx6Ho}{Video Summary -- https://youtu.be/qOTKRJMx6Ho}}.
As the pusher contact sticks to the object during a stable prehensile push, the object motion in the grasp can be inferred directly from the pusher motion. This provides full observability of the object pose in the grasp. Theoretically, we can track and control the pose of the object as accurately as the robot precision.

\section{Discussion}
\label{sec:discussion}

This paper demonstrates an effective use of sticking contact switch-overs for fast in-hand manipulation planning. We consider its application to prehensile pushing~\cite{ChavanDafle2015a,ChavanDafle2017} where a grasped object is manipulated using external pushes for which pusher contact sticks to the object. 

The main contribution of this paper is the formulation of the mechanics of stable prehensile pushing. We show that with the constraints of sticking mode at pusher contact, we can systematically fold down the dynamics of prehensile pushing, get rid of complementary constraints and write it as a constraint satisfaction problem. For manipulations in pinch grasp or parallel-jaw grasps it can further be efficiently solved with fast polyhedral geometry techniques.

To demonstrate the efficiency of the mechanics of stable prehensile pushing, we combined it with the T-RRT$^*$ based in-hand manipulation planning framework originally proposed in~\cite{ChavanDafle2017}.
Exploiting the features of T-RRT$^*$ and the dynamics formulation, the planner explores the configuration space of grasps on the object and generates a pushing strategy with minimal pusher switch-overs in a few seconds.

Experimental results validate the dynamics modeling of prehensile pushing. Pusher contacts indeed stick to the object for the pushing strategies planned for the variety manipulations we consider. 
This allows us to track the motion of the object in the grasp directly from the pusher motions which is the reflection of the robot motion in our case. 

Stable prehensile pushing empowers robots, even those with a simple gripper, to plan fast and observable in-hand manipulation strategies, one could say by equipping them with virtual actuators and virtual sensors.


\bibliographystyle{IEEEtranN} 
{\footnotesize \bibliography{ncd-icra18}}

\end{document}